\documentclass[11pt]{article}
\usepackage{acl2016}
\usepackage{times}
\usepackage{latexsym}

\usepackage{url}

\usepackage{times}
\usepackage{helvet}
\usepackage{courier}

\usepackage{graphicx}
\usepackage{latexsym}
\usepackage{url}
\usepackage{amsmath}
\usepackage{amsfonts}
\usepackage{multirow}

\title{On the Evaluation of Dialogue Systems with Next Utterance Classification }

\author{Ryan Lowe$^1$, Iulian V. Serban$^2$, Mike Noseworthy$^1$, Laurent Charlin$^{3}$\thanks{\;\;This work was primarily done while LC was at McGill University.} , Joelle Pineau$^1$\\
	    $^1$ School of Computer Science, McGill University\\
	    {\tt \{ryan.lowe, jpineau\}@cs.mcgill.ca,}\\ {\tt michael.noseworthy@mail.mcgill.ca}\\
	$^2$ DIRO, Universit{\'e} de Montr{\'e}al\\
	{\tt iulian.vlad.serban@umontreal.ca}\\
	$^3$ HEC Montr{\'e}al\\ 
	{\tt laurent.charlin@hec.ca}}

\aclfinalcopy
\date{}

\begin{document}

\maketitle

\begin{abstract}
An open challenge in constructing dialogue systems is developing methods for automatically learning dialogue strategies from large amounts of unlabelled data.  Recent work has proposed  Next-Utterance-Classification (NUC) as a surrogate task for building dialogue systems from text data.  In this paper we investigate the performance of humans on this task to validate the relevance of NUC as a method of evaluation.  Our results show three main findings:  (1) humans are able to correctly classify responses at a rate much better than chance, thus confirming that the task is feasible, (2) human performance levels vary across task domains (we consider 3 datasets) and expertise levels (novice vs experts), thus showing that a range of performance is possible on this type of task, (3) automated dialogue systems built using state-of-the-art machine learning methods have similar performance to the human novices, but worse than the experts, thus confirming the utility of this class of tasks for driving further research in automated dialogue systems.  

\end{abstract}

\section{Introduction}

Significant efforts have been made in recent years to develop computational methods for learning dialogue strategies offline from large amounts of text data. One of the challenges of this line of work is to develop methods to automatically evaluate, either directly or indirectly, models that are trained in this manner 
\cite{GalleyBSJAQMGD15,schatzmann2005quantitative}, without requiring human labels or human user experiments, which are time consuming and expensive.  The use of automatic tasks and metrics is one key issue in scaling the development of dialogue systems from small domain-specific systems, which require significant engineering, to general conversational agents \cite{pietquin2013survey}.

In this paper, we consider tasks and evaluation measures for what we call `unsupervised' dialogue systems, such as chatbots. These are in contrast to `supervised' dialogue systems, which we define as those that explicitly incorporate some supervised signal such as task completion or user satisfaction.
Unsupervised systems can be roughly separated into \textit{response generation} systems that attempt to produce a likely response given a conversational context, and \textit{retrieval-based} systems that attempt to select a response from a (possibly large) list of utterances in a corpus. While there has been significant work on building end-to-end response generation systems \cite{vinyals2015neural,shang2015neural,2015arXiv150704808S}, it has recently been shown that many of the automatic evaluation metrics used for such systems correlate poorly or not at all with human judgement of the generated responses \cite{liu2016not}.

Retrieval-based systems are of interest because they admit a natural evaluation metric, namely the recall and precision measures. First introduced for evaluating user simulations by Schatzmann et al. \shortcite{schatzmann2005quantitative}, such a framework has gained recent prominence for the evaluation of end-to-end dialogue systems~\cite{lowe2015ubuntu,kadlec2015improved,dodge2015}. These models are trained on the \textit{task} of selecting the correct response from a candidate list, which we call Next-Utterance-Classification (NUC, detailed in Section \ref{sec:nuc}), and are evaluated using the \textit{metric} of recall. 
NUC is useful for several reasons: 1) the performance (i.e. loss or error) is easy to compute automatically, 2) it is simple to adjust the difficulty of the task, 3) the task is interpretable and amenable to comparison with human performance, 4) it is an easier task compared to generative dialogue modeling, which is difficult for end-to-end systems \cite{sordoni2015aneural,2015arXiv150704808S}, and 5) models trained with NUC can be converted to dialogue systems by retrieving from the full corpus~\cite{liu2016not}. In this case, NUC additionally allows for making hard constraints on the allowable outputs of the system (to prevent offensive responses), and guarantees that the responses are fluent (because they were generated by humans). Thus, NUC can be thought of both as an \textit{intermediate task} that can be used to evaluate the ability of systems to understand natural language conversations, similar to the bAbI tasks for language understanding~\cite{weston2015towards}, and as a \textit{useful framework} for building chatbots. With the huge size of current dialogue datasets that contain millions of utterances~\cite{lowe2015ubuntu,Banchs:2012:MMD:2390665.2390716,Ritter:2010:UMT:1857999.1858019} and the increasing amount of natural language data, it is conceivable that retrieval-based systems will be able to have engaging conversations with humans.


However, despite the current work with NUC, there has been no verification of whether machine and human performance differ on this task. This cannot be assumed; it is possible that no significant gap exists between the two, as is the case with many current automatic response generation metrics~\cite{liu2016not}. Further, it is important to benchmark human performance on new tasks such as NUC to determine when research has outgrown their use.
In this paper, we consider to what extent NUC is achievable by humans, whether human performance varies according to expertise, and whether there is room for machine performance to improve (or has reached human performance already) and we should move to more complex conversational tasks.  
We performed a user study on three different datasets: 
the SubTle Corpus of movie dialogues~\cite{Banchs:2012:MMD:2390665.2390716}, the Twitter Corpus~\cite{Ritter:2010:UMT:1857999.1858019}, and the Ubuntu Dialogue Corpus~\cite{lowe2015ubuntu}. Since conversations in the Ubuntu Dialogue Corpus are highly technical, we recruit `expert' humans who are adept with the Ubuntu terminology, whom we compare with a state-of-the-art machine learning agent on all datasets. We find that there is indeed a significant separation between machine and expert human performance, suggesting that NUC is a useful intermediate task for measuring progress.

\section{Related Work}


Evaluation methods for supervised systems have been well studied. They include the PARADISE framework \cite{walker1997paradise}, and MeMo \cite{moller2006memo}, which include a measure of task completion. 
A more extensive overview of these metrics can be found in \cite{jokinen2009spoken}. We focus in this paper on unsupervised dialogue systems, for which proper evaluation is an open problem.


Recent evaluation metrics for unsupervised dialogue systems
include BLEU \cite{papineni2002bleu} and METEOR \cite{banerjee2005meteor}, which compare the similarity between response generated by the model, and the actual response of the participant, conditioned on some context of the conversation. Word perplexity, 
which computes a function of the probability of re-generating examples from the training corpus, is also used. 
However, such metrics have been shown to correlate very weakly with human judgement of the produced responses~\cite{liu2016not}.
They also suffer from several other drawbacks \cite{liu2016not}, including low scores, lack of interpretability, and inability to account for the vast space of acceptable outputs in natural conversation.  


\begin{figure}
    \centering
    \includegraphics[width=.95\linewidth]{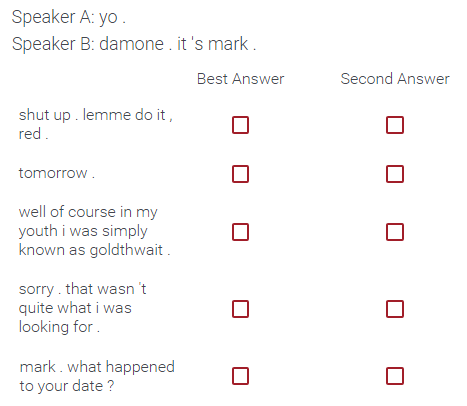}
    \caption{An example NUC question from the SubTle Corpus \cite{Banchs:2012:MMD:2390665.2390716}. }
    \label{fig:exquestion}
\end{figure}


\section{Technical Background on NUC}
\label{sec:nuc}
Our long-term goal is the development and deployment of artificial conversational agents. Recent deep neural architectures offer perhaps the most promising framework for tackling this problem. However training such architectures typically requires large amounts of conversation data from the target domain, and a way to automatically assess prediction errors. Next-Utterance-Classification (NUC, see Figure \ref{fig:exquestion}) is a \textit{task}, which is straightforward to evaluate, designed for training and validation of  dialogue systems. They are evaluated using the \textit{metric} of Recall@k, which we define in this section.

In NUC, a model or user, when presented with the context of a conversation and a (usually small) pre-defined list of responses, must select the most appropriate response from this list. This list \textit{includes the actual next response} of the conversation, which is the desired prediction of the model. The other entries, which act as false positives, are sampled from elsewhere in the corpus. Note that no assumptions are made regarding the number of utterances in the context: these can be fixed or sampled from arbitrary distributions. 
Performance on this task is easy to assess by measuring the success rate of picking the correct next response; more specifically, we measure Recall@k (R@k), which is the percentage of correct responses (i.e. the actual response of the conversation) that are found in the top $k$ responses with the highest rankings according to the model.  This task has gained some popularity recently for evaluating dialogue systems~\cite{lowe2015ubuntu,kadlec2015improved}. 

There are several attractive properties of this approach, as detailed in the introduction: the performance is easy to compute automatically, the task is interpretable and amenable to comparison with human performance, and it is easier than generative dialogue modeling. A particularly nice property is that 
one can adjust the difficulty of NUC by simply changing the number of false responses (from one response to the full corpus), or by altering the selection criteria of false responses (from randomly sampled to intentionally confusing). Indeed, as the number of false responses grows to encompass all natural language responses, the task becomes identical to response generation.



One potential limitation of the NUC approach is that, since the other candidate answers are sampled from elsewhere in the corpus, these may also represent reasonable responses given the context. Part of the contribution of this work is determining the significance of this limitation. 

\begin{table}[]
    \centering
    \footnotesize
    \begin{tabular}{|c|c|} \hline
         \multicolumn{2}{|c|}{What is your gender?} \\ 
         Male & 56.5\% \\
         Female & 44.5\% \\ \hline
         \multicolumn{2}{|c|}{What is your age?} \\ 
         18-20 & 3.4\% \\
         21-30 & 38.1\% \\
         31-40 & 33.3\% \\
         41-55 & 14.3\% \\
         55+ & 10.2\% \\ 
         \hline
         \multicolumn{2}{|c|}{How would you rate your fluency in English?} \\ 
         Beginner & 0\% \\
         Intermediate & 8.2\% \\
         Advanced & 6.8\% \\ 
         Fluent & 84.4\% \\ 
         \hline
         \multicolumn{2}{|c|}{What is your current level of education?} \\ 
         High school or less & 21.1\% \\
         Bachelor's & 60.5\% \\
         Master's & 13.6\% \\
         Doctorate or higher & 3.4\% \\ \hline
         \multicolumn{2}{|c|}{How would you rate your knowledge of Ubuntu?} \\ 
         I've never used it & 70.7\% \\
         Basic  & 21.8\% \\
         Intermediate & 5.4\% \\
         Expert & 2.7\% \\ \hline
    \end{tabular}
    \caption{Data on the 145 AMT participants.}
    \label{tab:demographics}
\end{table}

\section{Survey Methodology}

\subsection{Corpora}

We conducted our analysis on three corpora that have gained recent popularity for training dialogue systems. 
The SubTle Corpus \cite{Banchs:2012:MMD:2390665.2390716} consists of movie dialogues as extracted from subtitles, and includes turn-taking information indicating when each user has finished their turn. Unlike the larger OpenSubtitles\footnote{\url{http://www.opensubtitles.org}} dataset, the SubTle Corpus includes turn-taking information indicating when each user has finished their turn. 
The Twitter Corpus \cite{Ritter:2010:UMT:1857999.1858019} contains a large number of conversations between users on the microblogging platform Twitter. Finally, the Ubuntu Dialogue Corpus contains conversations extracted from IRC chat logs \cite{lowe2015ubuntu}.~\footnote{\url{http://irclogs.ubuntu.com}}
For more information on these datasets, we refer the reader to a recent survey on dialogue corpora \cite{serban2015survey}. 
We focus our attention on these as they cover a range of popular domains, and are among the largest available dialogue datasets, making them good candidates for building data-driven dialogue systems. 
Note that while the Ubuntu Corpus is most relevant to supervised systems, the NUC task still applies in this domain. Models that take semantic information into account (i.e., to solve the user's problem) can still be validated with NUC, as demonstrated in Lowe et al. \shortcite{lowe2015neural}.

A group of 145 paid participants were recruited through Amazon Mechanical Turk (AMT), 
a crowdsourcing platform for obtaining human participants for various studies.    Demographic data including age, level of education, and fluency of English were collected from the AMT participants, and is shown in Table \ref{tab:demographics}. 
An additional 8 volunteers were recruited from the student population in the computer science department at the author's institution.
\footnote{None of these participants were directly involved with this research project.}
This second group, referred to as ``Lab experts'', had significant exposure to technical terms prominent in the Ubuntu dataset; we hypothesized that this was an advantage in selecting responses for that corpus.

\subsection{Task description}

Each participant was asked to answer either 30 or 40 questions (mean=31.9).  To ensure a sufficient diversity of questions from each dataset, four versions of the survey with different questions were given to participants. For AMT respondents, the questions were approximately evenly distributed across the three datasets, while for the lab experts, half of the questions were related to Ubuntu and the remainder evenly split across Twitter and movies. Each question had 1 correct response, and 4 false responses drawn uniformly at random from elsewhere in the (same) corpus. An example question can be seen in Figure \ref{fig:exquestion}.
Participants had a time limit of 40 minutes. 

Conversations were extracted to form NUC conversation-response pairs as described in Sec.~\ref{sec:nuc}.  The number of utterances in the context were sampled according to the procedure in \cite{lowe2015ubuntu}, with a maximum context length of 6 turns --- this was done for both the human trials and ANN model. All conversations were pre-processed in order to anonymize the utterances. For the Twitter conversations, this was extended to replacing all user mentions (words beginning with @) throughout the utterance with a placeholder `@user' symbol, as these are often repeated in a conversation. Hashtags were not removed, as these are often used in the main body of tweets, and many tweets are illegible without them. 
Conversations were edited or pruned to remove offensive language according to ethical guidelines.

\begin{table*}[]
    \small
    \centering
    \begin{tabular}{|l|c|c c|c c|c c|}
    \hline
    \multirow{2}{*}{ } & \multirow{1}{*}{Number} &
    \multicolumn{2}{|c|}{\textbf{Movie Corpus}} &
    \multicolumn{2}{|c|}{\textbf{Twitter Corpus}}&
    \multicolumn{2}{|c|}{\textbf{Ubuntu Corpus}} \\ 
      & of Users & R@1 & R@2 & R@1 & R@2 & R@1 & R@2  \\ \hline
       AMT non-& \multirow{2}{*}{135} & \multirow{2}{*}{65.9 $\pm$ 2.4\% }& \multirow{2}{*}{79.8 $\pm$ 2.1\%} & \multirow{2}{*}{74.1 $\pm$ 2.3\%} & \multirow{2}{*}{82.3 $\pm$ 2.0\%} & \multirow{2}{*}{52.9 $\pm$ 2.7\%} & \multirow{2}{*}{69.4 $\pm$ 2.5\%} \\ 
        \hspace{1mm} experts & & & & & & & \\
       AMT experts & 10 & --- & --- & --- & --- & 52.0 $\pm$ 9.8\% & 63.0 $\pm$ 9.5\% \\
       Lab experts& 8 &  69.7 $\pm$ 10\% & 94.0 $\pm$ 5.2\%$^*$ &  88.4 $\pm$ 7.0\% & 98.4 $\pm$ 2.7\%$^*$ &  83.8 $\pm$ 8.1\% & 87.8 $\pm$ 7.2\% \\ \hline 
       ANN model & \multirow{3}{*}{machine}   & \multirow{3}{*}{50.6\%} & \multirow{3}{*}{74.9\%}  & \multirow{3}{*}{66.9\%} & \multirow{3}{*}{89.6\%} & \multirow{3}{*}{66.2\%} & \multirow{3}{*}{83.7\%} \\ 
       (Lowe et al., & & & & & & &\\
       2015a) & & & & & & & \\
       \hline
    \end{tabular}
    \caption{Average results on each corpus. `Number of Users' indicates the number of respondents for each category. `AMT experts' and `AMT non-experts' are combined for the Movie and Twitter corpora. 95\% confidence intervals are calculated using the normal approximation, which assumes subjects answer each question independently of other examples and subjects. Starred (*) results indicate a poor approximation of the confidence interval due to high scores with small sample size, according to the rule of thumb by Brown et al. \shortcite{brown2001interval}.}
    \label{tab:results}
\end{table*}

\subsection{ANN model}

In order to compare human results with a strong artificial neural network (ANN) model, we use the dual encoder (DE) model from Lowe et al. \shortcite{lowe2015ubuntu}. This model uses recurrent neural networks (RNNs) with long-short term memory (LSTM) units \cite{hochreiter1997long} to encode the context $c$ of the conversation, and a candidate response $r$. More precisely, at each time step, a word $x_t$ is input into the LSTM, and its hidden state is updated according to: $h_t = f(W_h h_{t-1} + W_x x_t)$, where $W$ are weight matrices, and $f(\cdot)$ is some non-linear activation function. After all $T$ words have been processed, the final hidden state $h_T$ can be considered a vector representation of the input sequence. 

To determine the probability that a response $r$ is the actual next response to some context $c$, the model computes a weighted dot product between the vector representations $\mathbf{c}$, $\mathbf{r} \in \mathbb{R}^d$ of the context and response, respectively:
$$
P(r \text{ is correct response}) = \sigma (\mathbf{c}^\top M \mathbf{r})
$$
where $M$ is a matrix of learned parameters, and $\sigma$ is the sigmoid function. The model is trained to minimize the cross-entropy error of context-response pairs. For training the authors randomly sample negative examples.

The DE model is close to state-of-the-art for neural network dialogue models on the Ubuntu Dialogue Corpus; we obtained further results on the Movie and Twitter corpora in order to facilitate comparison with humans. For further model implementation details, see Lowe et al. \shortcite{lowe2015ubuntu}.

\section{Results}
As we can see from Table \ref{tab:demographics}, the AMT participants are mostly young adults, fluent in English with some undergraduate education. The split across genders is approximately equal, and the majority of respondents had never used Ubuntu before.

Table \ref{tab:results} shows the NUC results on each corpus. The human results are separated into AMT non-experts, consisting of paid respondents who have `Beginner' or no knowledge of Ubuntu terminology; AMT experts, who claimed to have `Intermediate' or `Advanced' knowledge of Ubuntu; and Lab experts, who are the non-paid respondents with Ubuntu experience and university-level computer science training. 
We also presents results on the same task for a state-of-the-art artificial neural network (ANN) dialogue model (see \cite{lowe2015ubuntu} for implementation details).

We first observe that subjects perform above chance level ($20\%$ for R@1) on all domains, thus the task is doable for humans. Second we observe difference in performances between the three domains. The Twitter dataset appears to have the best predictability, with a Recall@1 approximately 8\% points higher than for the movie dialogues for AMT workers, and 18\% higher for lab experts. Rather than attributing this to greater familiarity with Twitter than movies, it seems more likely that it is because movie utterances are often short, generic (e.g.\@ contain few topic-related words), and lack proper context (e.g., video cues and the movie's story). Conversely, tweets are typically more specific, and successive tweets may have common hashtags.

As expected, untrained respondents scored lowest on the Ubuntu dataset, as it contains the most difficult language with often unfamiliar terminology. Further, since the domain is narrow, randomly drawn false responses could be more likely to resemble the actual next response, especially to someone unfamiliar with Ubuntu terminology. 
We also observe that the ANN model achieves similar performance to the paid human respondents from AMT. However, the model is still significantly behind the lab experts for Recall@1.

An interesting note is that there is very little difference between the paid AMT non-experts and AMT experts on Ubuntu.
This suggests that the participants do not provide accurate self-rating of expertise, either intentionally or not.
We also found that lab experts took on average approximately 50\% more time to complete the survey than paid testers; this is reflected in the results, where the lab experts score 30\% higher on the Ubuntu Corpus, and even 5-10\% higher on the non-technical Movie and Twitter corpora. While we included attention check questions to ensure the quality of responses,\footnote{Only the respondents who passed all attention checks were counted in the survey.} this reflects poorly on the ability of crowdsourced workers  
to answer technical questions, even if they self-identify as being adept with the technology.

\section{Discussion}


Our results demonstrate that humans outperform current dialogue models on the task of Next-Utterance-Classification, indicating that there is plenty of room for improvement for these models to better understand the nature of human dialogue. While our results suggest that NUC is a useful task, it is by no means sufficient; we strongly advocate for automatically evaluating dialogue systems with as many relevant metrics as possible. Further research should be conducted into finding metrics or tasks which accurately reflect human judgement for the evaluation of dialogue systems.







\paragraph{Acknowledgements}

The authors gratefully acknowledge financial support  for  this  work  by  the  Samsung  Advanced
Institute  of  Technology  (SAIT)  and  the  Natural Sciences  and  Engineering  Research  Council  of Canada  (NSERC).

\bibliography{acl2016}
\bibliographystyle{acl2016}

\end{document}